\documentclass{article}
\usepackage{spconf,amsmath,graphicx}
\usepackage[utf8]{inputenc} 
\usepackage[english]{babel}
\usepackage[T1]{fontenc}    
\usepackage{hyperref}       
\usepackage{url}            
\usepackage{booktabs}       
\usepackage{amsfonts}       
\usepackage{nicefrac}       
\usepackage{microtype}      
\usepackage{bm} 
\usepackage[square,sort,comma,numbers]{natbib}
\usepackage{amsmath}
\usepackage{wrapfig}

\usepackage[table]{xcolor}
\usepackage{graphicx} 
\usepackage{algorithm}
\usepackage{amsmath}
\usepackage{amssymb}
\usepackage{amsfonts}
\usepackage{dsfont}
\usepackage{xspace}
\usepackage{xr}
\usepackage{multirow}
\usepackage{multicol}
\usepackage{siunitx}
\usepackage{afterpage}
\usepackage{enumerate}
\usepackage{booktabs}
\usepackage{natbib}
\usepackage{soul}
\usepackage{mathtools}
\usepackage{amsthm}

\usepackage{booktabs}
\usepackage{bm}
\usepackage{subcaption}
\usepackage{algpseudocode}

\theoremstyle{definition}
\newtheorem{theorem}{Theorem}

\newcommand{\eqdef}{\mathrel{\overset{\mathrm{def}}{=}}}
\DeclareMathOperator*{\argmin}{arg\,min}

\title{Investigating Generalization in Neural Networks under\\ Optimally Evolved Training Perturbations}
\name{\begin{tabular}{c} Subhajit Chaudhury$^{\star \dagger}$ \qquad Toshihiko Yamasaki$^{\star}$ \\ \{subhajit, yamasaki\}@hal.t.u-tokyo.ac.jp \end{tabular}}
\address{$^{\star}$ The University of Tokyo \qquad $^{\dagger}$IBM Research AI - Tokyo}

\begin{document}
	\maketitle
	\begin{abstract}
	In this paper, we study the generalization properties of neural networks under input perturbations and show that minimal training data corruption by a few pixel modifications can cause drastic overfitting. We propose an evolutionary algorithm to search for optimal pixel perturbations using novel cost function inspired from literature in domain adaptation that explicitly maximizes the generalization gap and domain divergence between clean and corrupted images. Our method outperforms previous pixel-based data distribution shift methods on state-of-the-art Convolutional Neural Networks (CNNs) architectures. Interestingly, we find that the choice of optimization plays an important role in generalization robustness due to the empirical observation that SGD is resilient to such training data corruption unlike adaptive optimization techniques~(ADAM). Our Source code is available at \url{https://github.com/subhajitchaudhury/evo-shift}.
	\end{abstract}
	\vspace{-0.1cm}
	\section{Introduction}
	\label{sec:intro}
	Deep learning has shown notable empirical success in various application areas. Typically, in an over-parametrized setting with a highly non-convex loss surface, classical learning theory~\cite{vapnik2013nature} predicts that deep neural networks should have a high out-of-sample error because the solution is likely to get stuck at a local minimum. Nonetheless, deep neural networks appear to generalize well even in small data regimes. Numerous recent works have sought to explain generalization in neural networks. Zhang et al.~\cite{zhang2016understanding} showed that neural networks can fit random noise and labels, thus refuting the finite sample expressivity argument. Another view~\cite{keskar2016large} as to why neural networks generalize well, studies the loss surface geometry around the learned parameter and shows that sharper minima solutions tend to generalize poorly compared to flatter minima which were contested by Dinh et al.~\cite{dinh2017sharp}. Some recent research~\cite{keskar2017improving, wilson2017marginal} also demonstrates that vanilla SGD optimization has better generalization ability than adaptive optimization methods. 
		\vspace{-0.1cm}
\begin{figure}[tb]
	\label{fig:opening}
	\centering
	\includegraphics[width=0.99\linewidth]{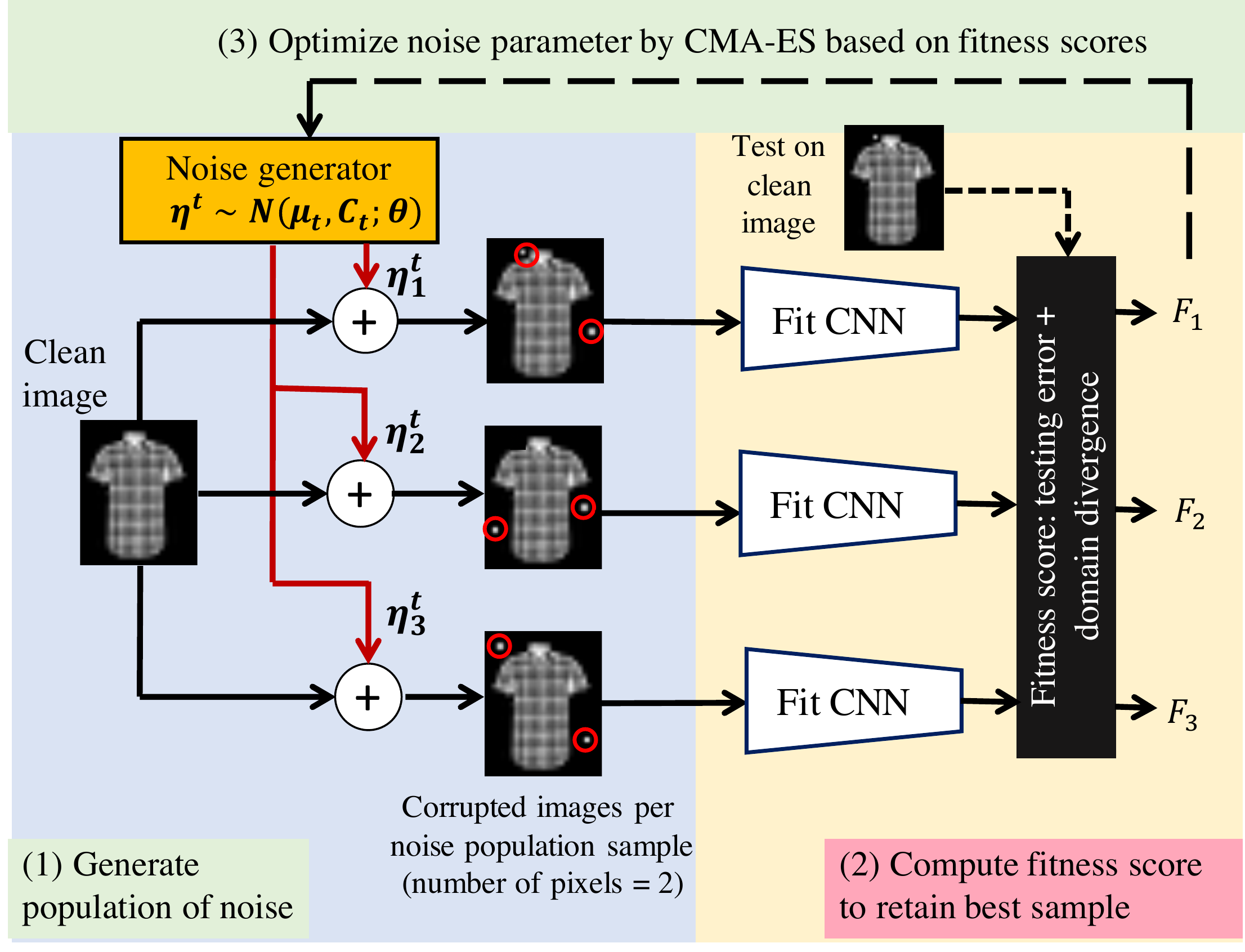}
	\caption{Overview of our proposed noise optimization algorithm}
\end{figure}   
	
	Our method is similar to \textit{Adversarial Distribution Shift}~(ADS) presented in \cite{jacobsen2018excessive} where benign perturbations are added to the training data causing neural networks to learn task-irrelevant features.  Specifically, \cite{jacobsen2018excessive} studied the effect of single-pixel perturbations on MNIST training images on clean test performance. Data poisoning attacks~\cite{biggio2012poisoning, shafahi2018poison, steinhardt2017certified} are also related to such an approach where the adversary injects a few malicious samples in the training data to cause incorrect classification~(typically targeted) during inference. Tanay et al.~\cite{tanay2018built} showed that neural network models can be made almost arbitrarily sensitive to a single-pixel while maintaining identical test performance between models. However, poisoning methods~\cite{munoz2017towards, shafahi2018poison, koh2017understanding} usually modify some part of the decision boundary by adding malicious training samples for targeted misclassifications, which is different from our approach of optimal ADS. Moreover, our motivation in this work is to analyze how optimization methods, specifically adaptive and non-adaptive algorithms, contribute to generalization robustness which is different from the typical objective of data poisoning methods. 
	
	In this paper, we find optimal training ADS that cause a high generalization gap between corrupted training and clean images during inference while limiting the attack to a few pixels only. The overview of our method is shown in Figure~1.
	Our contribution in this paper is two-fold. Firstly, we propose a novel fitness function for the CMA-ES algorithm to find optimal pixel disturbance, using domain adaptation theory. Our method outperforms previous heuristic ADS method presented in \cite{jacobsen2018excessive}. Secondly, our analysis reveals that the choice of optimization technique plays an important role in generalization robustness. Specifically, vanilla SGD is found to be surprisingly resilient against training sample perturbations compared to adaptive optimization methods like ADAM, which calls into question the effectiveness of such popular adaptive optimization methods towards generalization robustness.

		\vspace{-0.1cm}
	\section{Problem Setup}
	\label{sect:problem}
	We consider a multi-class classification task with input space $X \in \mathbb{R}^N$ and label space $Y = \{1,...,N_c\}$. The true data distribution is given as, $ S = \{ \bm{x}_i, y_i \}_{i=1}^n \sim \mathcal{D}_S $. Our goal is to train a classifier on a perturbed version of the true data samples such that the empirical risk~(or test error) on the true uncorrupted samples is maximized. Considering that for each sample in $S$, we can draw class-wise input perturbations, $\delta = \{ \bm{\eta}_i \}_{i=1}^{N_c} \sim {N(\bm{m}, \bm{\Sigma})}$, parameterized by the mean $\bm{m}$ and covariance matrix $\bm{\Sigma}$, which are added to the true samples, $\bm{x}^p_i=\bm{x}_i+\bm{\eta}_{y_i}$, where noise encoding each class information is added to training images. The joint distribution of the perturbed data, constructed by assigning labels of the true samples to the corresponding perturbed samples, given as $ P = \{ \bm{x}^p_i, y^p_i \}_{i=1}^n \sim \mathcal{D}_P $. In this paper, we work with image inputs and perturb a few pixels to analyze generalization sensitivity to small changes in training inputs. 
	
	Let us define a classifier function $h : X \rightarrow Y$ from a hypothesis space $\mathcal{H}$. The corresponding empirical risk on samples drawn from a distribution $\mathcal{D}$ is defined as, $R_{\mathcal{D}}(h) \eqdef \mathbb{E}_{(\bm{x}, y) \sim \mathcal{D}}\big( I [h(\bm{x}) \ne y] \big)$, which signifies the error on the samples drawn from $\mathcal{D}$. Our objective is to find optimal perturbation parameter that increases the empirical risk on the clean samples while minimizing it on the corrupted samples, thus compromising generalization in neural networks, given as
	
	\begin{equation}
	\label{eq:obj}
    \max_{\bm{m}, \mathbf{\Sigma}} \bigg( R_{\mathcal{D}_S}(h^*) - R_{\mathcal{D}_P}(h^*) \bigg) \; s.t.\; h^* = \argmin_{h \in \mathcal{H}} R_{\mathcal{D}_P}(h).
	\end{equation}
	
	The above objective finds optimal perturbation parameter that increases the empirical risk on the clean samples while minimizing it on the corrupted samples, thus compromising generalization in neural networks.
	
	\section{Maximum Domain Divergence based Evolutionary Strategy (MDD-ES)} 
	
	The objective function in Equation \ref{eq:obj} requires a nested minimization for classifier training and empirical risk maximization for optimal noise search. This presents difficulty in using standard gradient-based optimization methods for searching the optimal pixel perturbations. Therefore, we use a black-box optimization technique, specifically Covariance Matrix Adaptation Evolution Strategy~(CMA-ES)~\cite{hansen2016cma}, which has been shown to work well in high-dimensional problems~\cite{ha2018recurrent}. However, simply using empirical risk~(generalization gap) measure on clean samples as a fitness score might require more generations for convergence. However, each generation of the CMA-ES is computationally expensive (due to multiple CNN training rounds). Therefore, we propose a novel fitness score inspired by the domain divergence literature that provides an additional signal for convergence, leading to improved noise optimization properties from fewer generations.
	
	\subsection{Measuring Domain-Divergence}
	Considering a domain $\mathcal{X}$ and a collection of subsets of $\mathcal{X}$ as $\mathcal{A}$. Given two domain distributions $D_S$ and $D_T$ over~$\mathcal{X}$, and a hypothesis class~$\mathcal{H}$, Shai et al.~\cite{ben2007analysis, ganin2016domain} showed that domain divergence~($\mathcal{H}$-divergence) for the hypothesis space of linear classifiers can be approximately computed by the empirical $\mathcal{H}$-divergence from samples $\bm{x}^s_i \sim \tilde{D}_S$ and $\bm{x}^t_i \sim \tilde{D}_T$ as,
	
	\begin{equation} \label{eq:Hdiv_empirical}
	\begin{split}
	\hat{d}_{\mathcal{H}}(S,T) \eqdef 2\Bigg( & 1 - \min_{h\in\mathcal{H}} \bigg[
	\frac{1}{n} \sum_{i=1}^n I[h(\bm{x}^s_i)\!=\!0] \\
	& + \frac{1}{n'} \sum_{i=n+1}^{N} I[h(\bm{x}^t_i)\!=\!1]
	\bigg] \Bigg)\,,
	\end{split}
	\end{equation}
	
		\noindent where $n$ samples from the source domain and $n'$ samples from the target domain is drawn. The proxy $\mathcal{A}$-distance is computed as, $\hat{d}_{\mathcal{A}} = 2(1-2\epsilon)$ according to ~\cite{ben2007analysis}, where $\epsilon$ is the discriminator error.

	\subsection{Bound on Target Risk}
	
	We are interested in finding a bound of the target empirical risk obtained by learning a classifier of the source samples. Shai et al.~(and later used by Ganin et al.~\cite{ben2007analysis, ben2010theory, ganin2016domain}) showed that the bound on target risk can be computed in terms of the proxy $\mathcal{A}$-distance defined above, as follows,

	\begin{theorem}
		\label{thm:RDT_bound}
		Considering $\mathcal{H}$ be a hypothesis class of VC dimension $d$, for $n$ samples $S\sim (\tilde{D}_S)^n$ and $T\tilde (\hat{D}_T)^{n}$, then with probability $1-\delta$ over the choice of samples, for every $h\in\mathcal{H}$:
		\begin{equation}
		\begin{split}
		\hat{R}_T(h) & \leq  \hat{R}_S(h) +  \sqrt{\frac{4}{n}\left( d \log\tfrac{2e\, n}{d}+  \log\tfrac{4}{\delta}\right) } \\
		& +\hat{d}_{\mathcal{H}}(S,T) + 4\sqrt{ \frac{1}{n}\left(d \log\tfrac{2 n}{d}+  \log\tfrac{4}{\delta}\right) }
		+ \beta ,
		\end{split}
		\end{equation}
	\end{theorem}
	
	\noindent with $\beta \geq {\displaystyle\inf_{h^*\in\mathcal{H}}} \left[ R_S(h^*) + R_T(h^*) \right]$\, and $\hat{R}_S(h) $ is the empirical source risk. 
	
	Given a fixed hypothesis space, we observe that increasing the $\mathcal{H}$-divergence between the two domains would make the above bound loose. Since we are interested in maximizing the target risk, pixel perturbations that increase the $\mathcal{H}$-divergence between corrupted and clean data would be more likely to fool the neural network. We use this insight to craft a fitness score that favors solutions with high domain divergence between the clean and perturbed distributions.
	
	\vspace{-0.1cm}
	\subsection{Proposed Fitness Score based CMA-ES Optimization}
	Using the insights developed in the previous section, we propose MDD-ES algorithm that utilizes a fitness score measuring, (i) semantic mismatch score, (ii) domain divergence score. Given training data, $(x,y) \sim \mathcal{D}$, and initial CMA-ES parameters, $\bm{m}_0, \mathbf{\Sigma}_0, \sigma_0$, we sample a population of noise for each generation, $\{\delta_j\}_{j=1}^{\lambda} \sim {N(\bm{m}_t, \mathbf{\Sigma}_t)}$. For each sample in the current generation $t$, we obtain the optimal weights, $\bm{\theta}^*$, by training a CNN~($F_\theta^j$) from scratch on the corrupted training samples $\{x+\delta_j\}$. We compute the semantic mismatch score for the $j^{th}$ noise sample as 
	$\mathcal{F}^j_m= \mathbb{E}_{(x,y) \sim \mathcal{D}} \big[ l_{CE}(F_\theta^j(x+\delta_j),y) - l_{CE}(F_\theta^j(x),y) \big]$, where $ l_{CE}$ is the cross-entropy loss. This score encourages high loss of generalization between clean and corrupted samples drawn from the training distribution. To obtain the domain divergence score, we train a discriminator with corrupted samples as label $0$ and clean samples as label $1$. The domain divergence score is computed as, ${F}^j_d = (1-2\epsilon)$, where $\epsilon$ is the error of the trained discriminator. The overall fitness score for the CMA-ES algorithm is computed as the combination of above score, $\mathcal{F}_j$ = $\mathcal{F}^j_m + \mathcal{F}^j_d$.  After each generation, the sampling parameters are updated by the CMA-ES algorithm to favor the pixel perturbations corresponding to the top-performing fitness scores, $\bm{m}_{t+1}, \mathbf{\Sigma}_{t+1}, \sigma_{t+1} = \text{CMA-ES}(\bm{m}_t, \bm{\Sigma}_t, \sigma_t, \mathcal{F}_j)$. We refer the reader to the original paper~\cite{hansen2016cma} for details on the CMA-ES update algorithm. The best performing fitness score across all generations is chosen as the optimal pixel perturbations, $\delta^*$. It must be noted that no samples from the testing data was used in the training phase for optimizing the noise generator parameters. During testing, we train with the optimally corrupted training data and perform inference on clean test data. 
	
%

		\vspace{-0.1cm}
	
	\section{Experimental Results}
	We evaluated our method on four datasets: MNIST, Fashion-MNIST, SVHN cropped $32 \times 32$ images, and CIFAR10 images. The perturbed MNIST images for $N_p=1$ are shown in Figure~\ref{fig:intro}~(a). Learning perturbations by evolution involves multiple training rounds in each generation. We used two custom CNN models as underlying models in the evolutionary learning stage: GrayNet~(24C3-P-48C3-P-256FC-10S),  for MNIST, Fashion-MNIST and ColorNet~(32C3-32C3-P-64C3-64C3-P-128C3-128C3-P-512FC-10S) for CIFAR10, SVHN dataset. We use four settings of number of pixel perturbation, $N_p = \{1, 2, 5, 10\}$. 
	
	\vspace{-0.1cm}
	
	\subsection{Learning Curves for Perturbation Optimization}
	
	\begin{figure}[tb]
		\centering
		\includegraphics[width=0.99\linewidth]{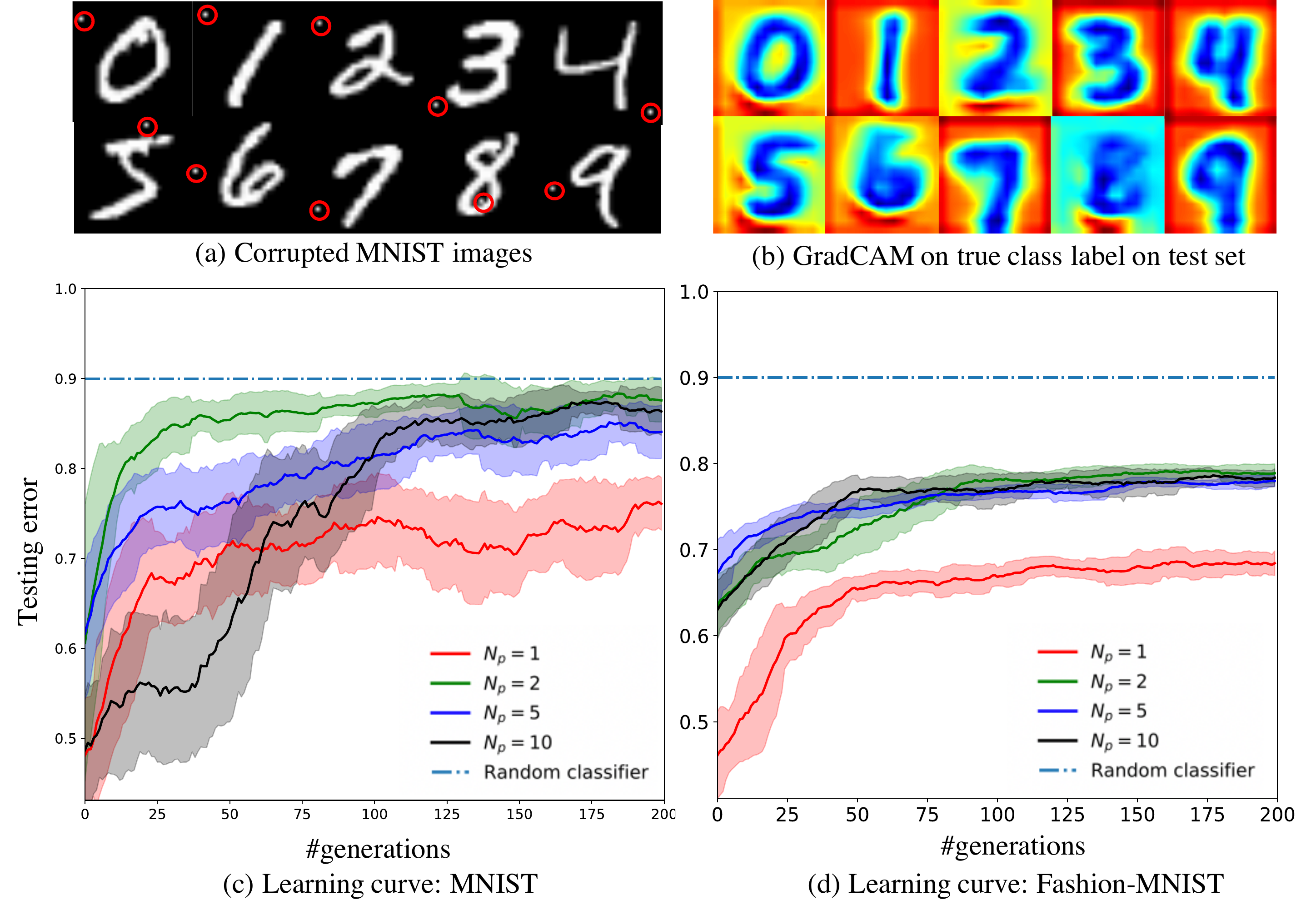}
		\caption{(a) Highlighting learned single pixel perturbations on MNIST images, (b) GradCAM visualization of the last Conv layer for $N_p = 1$. Dominant gradient distribution is on the background. Learning curve with increasing generations of CMA-ES is shown for (c) MNIST and (d) Fashion-MNIST}\label{fig:intro}
	\end{figure}    
	
	\begin{table}[tb]
		\centering
		\begin{tabular}{lcccc}
			\toprule
			\textbf{Method}  & \textbf{ResNet-20} & \textbf{ResNet-32} & \textbf{DenseNet-40} \\
			\midrule
			\textbf{SVHN}(clean) & ${93.5} \pm {0.9}$  & ${92.8} \pm {1.0}$  & ${92.3} \pm {1.2}$    \\
			{$N_p=1$~[Baseline]} & ${30.3} \pm {8.6}$  & ${41.4} \pm {8.9}$  & ${36.3} \pm {4.2}$ \\
			{$N_p=1$~\cite{jacobsen2018excessive}} & $91.8 \pm {0.2}$  & ${90.9} \pm {1.8}$  & ${91.0} \pm {0.4}$ \\
			{$N_p=1$, [ours]} & $31.3 \pm 6.3$  & $37.2 \pm 10.4$  & $32.1 \pm 9.4$   \\
			{$N_p=2$, [ours]} & $14.9 \pm 2.4$  & $18.4 \pm 3.8$  & $18.8 \pm 4.7$   \\
			{$N_p=5$, [ours]} & $\textbf{9.3} \pm \textbf{0.9}$  & $\textbf{11.0} \pm \textbf{0.3}$  & $\textbf{16.1} \pm \textbf{8.4}$  \\
			\bottomrule
		\end{tabular}
		\caption{Showing testing accuracy~(in \%) on clean test samples, trained on optimally perturbed samples with DA for $30$ epochs on SVHN dataset. Experiments are repeated 3 times.}
		\label{table:comparison}
	\end{table}
	
	We examine test error with increasing generations of our proposed algorithm as shown in Figure~\ref{fig:intro}~(c) and Figure~\ref{fig:intro}~(d) for MNIST and Fashion-MNIST datasets respectively. Test error is seen to grow as the evolutionary optimization advances indicating the soundness of our proposed optimization strategy. Additionally, we visualize the mean GradCAM distribution of $100$ images per class from the testing dataset corresponding to the true class label for MNIST dataset in Figure~\ref{fig:intro}~(b), which reveals that the CAM distribution shifts its density to non-salient background ROI in the image, thus learning non-discriminative features that do not generalize well. This might explain the drop in testing accuracy with increasing epochs.
	
	\subsection{Comparison to Prior Methods}
	As a baseline for our task, we choose a uniformly sampled spatial distribution of pixel perturbation, which is the starting point of the CMA-ES algorithm. Our method consistently outperforms both the baseline method and Jacobsen et al.~\cite{jacobsen2018excessive} on the metric of test error on the clean test set, for all the datasets as shown in Table~\ref{table:comparison}. Our method shows superior performance compared to \cite{jacobsen2018excessive} because we perform optimization to search for the best corruption pattern whereas \cite{jacobsen2018excessive} uses heuristic pixel perturbations on the left-most column of the input image to encode class specific information. The baseline method outperforms Jacobsen et al.~\cite{jacobsen2018excessive} due to data augmentation.
	
	\begin{figure}[t]
		\centering
		\includegraphics[width=0.99\linewidth]{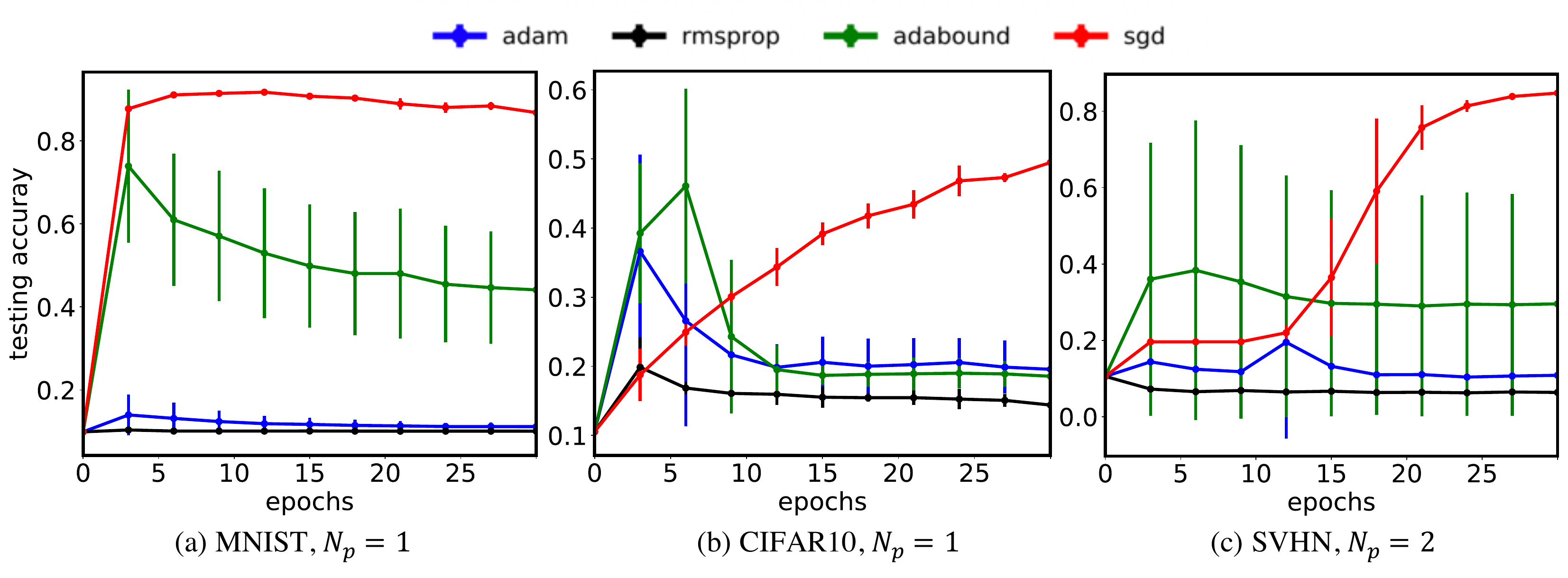}
		\caption{Testing accuracy using various optimization strategies under proposed perturbation shows SGD consistently performs better than adaptive optimization techniques. Each experiment was performed 5 times and one std dev. error is shown.}\label{fig:optimplot}
		\label{fig:optim}
	\end{figure}
	
	\subsection{Adaptivity can Overfit to Training Perturbations}
	
	High out-of-sample error is generally attributed to poor convergence of the neural network parameters to an unfavorable local minimum. By examining the robustness of well-known optimization strategies to our proposed pixel perturbation algorithm, we wish to study if a certain algorithm is more liable to memorizing small perturbations while ignoring other salient statistical patterns in the training data. To this end, we trained CNN models on our proposed optimal ADS data using ADAM~\cite{kingma2014adam}, SGD, RMSProp~\cite{tieleman2017divide}, and Adabound~\cite{luo2019adaptive} optimization. The results are shown in Figure~\ref{fig:optimplot}.
	
	Wilson et al.~\cite{wilson2017marginal} showed that adaptive methods are affected by spurious features that do not contribute to out-of-sample generalization by crafting a smart artificial linear regression example. 
	Our method is an extension of such methods for automatic creation of spurious examples that scale to arbitrarily sized datasets by evolutionary strategies. Figure~\ref{fig:optimplot} reveals that ADAM and RMSProp show prohibitively low testing accuracy for all cases while vanilla SGD is surprisingly resilient to such perturbations showing better out-of-sample performance consistently for all the datasets. Adabound uses strategies from both SGD and Adam, thus showing intermediate performance. Thus, adaptive methods overfit to training perturbations while vanilla SGD is considerably robust to such changes.
	
	\begin{figure}[tb]
		\centering
		\includegraphics[width=0.99\linewidth]{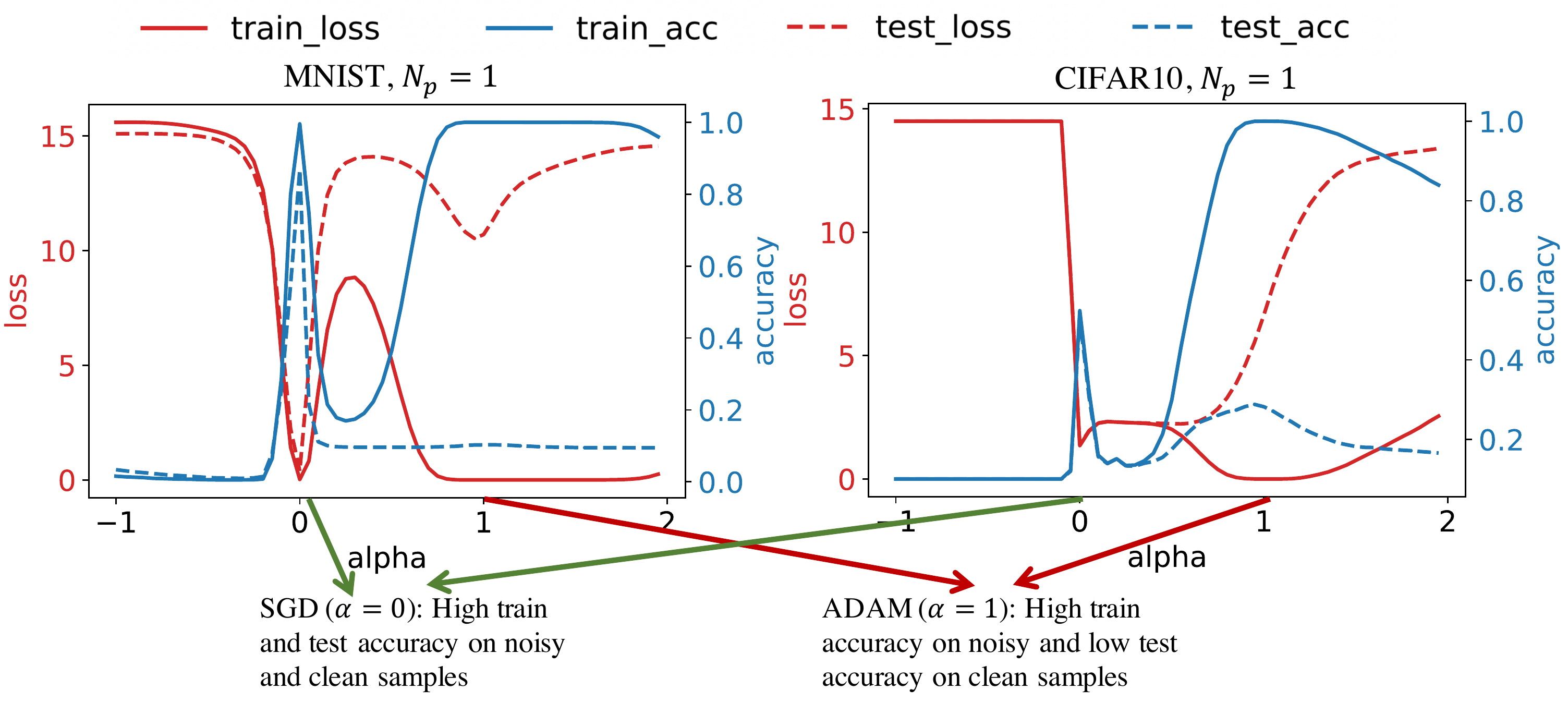}
		\caption{Interpolating loss surface from SGD~($\alpha=0$) to ADAM~($\alpha=1$) weights. The loss surface around SGD parameter is sharper however has better generalization.}\label{fig:lossplot}
	\end{figure}

	Due to the input data corruption, the loss manifold changes to favor solutions that overfit to the spurious perturbation features. Our intuition is that adaptive methods adjust an algorithm to the geometry of the data~\cite{wilson2017marginal} and thus overfits to such spurious features. In contrast, SGD's optimization strategy does not depend on the data, but it uses the $l_2$ geometry inherent to the parameter space. Thus it performs better than adaptive optimization algorithms. 
	
	\textbf{Loss surface} : Keskar et al.~\cite{ keskar2016large, hochreiter1997flat} claimed that flatter minima solutions generalize better compared to its sharper counterparts. To investigate this phenomenon, we visualize the loss surface around the learned parameters by interpolating between the weights obtained from SGD and ADAM optimization following the strategy by Goodfellow et al.~\cite{goodfellow2014qualitatively}. We plot the loss function values and train/test accuracies at intermediate intervals given as $\bm{w}_{\alpha}=\alpha \bm{w}_{\text{ADAM}} + (1 - \alpha )\bm{w}_{\text{SGD}}$ as shown in Figure~\ref{fig:lossplot}. Interestingly, we find that SGD finds sharper minima solutions where both test and train loss are low~($\alpha=0$) compared to ADAM, where the train loss exhibits are more flatter geometry~($\alpha=1$). This pattern is repeatedly visible for all datasets suggesting that sharpness of minima does not guarantee a solution that has better generalization robustness to training perturbations, which is along the same line of argument as claimed by Dinh et al.~\cite{dinh2017sharp}.
	
	\section{Conclusion}
	We present a population-based evolutionary strategy using a novel fitness score to search for pixel perturbations that explicitly maximize domain divergence and generalization gap. Our method incrementally fools the neural networks with each passing generation suggesting the existence of certain vulnerable spatial locations on input images. Our analysis reveals that a proper selection of neural network optimization is paramount to good generalization. We find that vanilla SGD performs significantly better than adaptive optimization methods in ignoring spurious training features that do not contribute to out-of-sample generalization. Our analysis of loss surface, reveals that in spite of good generalization performance SGD finds sharper minima solutions than ADAM. It might be tempting to conclude that sharper minima solutions are more robust to input perturbation overfitting however more analysis is required in this direction. We believe this work will fuel further research into understanding the generalization properties of deep learning optimization in the presence of input noise.
	
	\bibliographystyle{IEEEbib}
	\bibliography{strings}

\begin{thebibliography}{10}

\bibitem{vapnik2013nature}
Vladimir Vapnik,
\newblock {\em The nature of statistical learning theory},
\newblock Springer science \& business media, 2013.

\bibitem{zhang2016understanding}
Chiyuan Zhang, Samy Bengio, Moritz Hardt, Benjamin Recht, and Oriol Vinyals,
\newblock ``Understanding deep learning requires rethinking generalization,''
\newblock {\em arXiv preprint arXiv:1611.03530}, 2016.

\bibitem{keskar2016large}
Nitish~Shirish Keskar, Dheevatsa Mudigere, Jorge Nocedal, Mikhail Smelyanskiy,
  and Ping Tak~Peter Tang,
\newblock ``On large-batch training for deep learning: Generalization gap and
  sharp minima,''
\newblock {\em arXiv preprint arXiv:1609.04836}, 2016.

\bibitem{dinh2017sharp}
Laurent Dinh, Razvan Pascanu, Samy Bengio, and Yoshua Bengio,
\newblock ``Sharp minima can generalize for deep nets,''
\newblock in {\em Proceedings of the 34th International Conference on Machine
  Learning-Volume 70}. JMLR. org, 2017, pp. 1019--1028.

\bibitem{keskar2017improving}
Nitish~Shirish Keskar and Richard Socher,
\newblock ``Improving generalization performance by switching from adam to
  sgd,''
\newblock {\em arXiv preprint arXiv:1712.07628}, 2017.

\bibitem{wilson2017marginal}
Ashia~C Wilson, Rebecca Roelofs, Mitchell Stern, Nati Srebro, and Benjamin
  Recht,
\newblock ``The marginal value of adaptive gradient methods in machine
  learning,''
\newblock in {\em Advances in Neural Information Processing Systems}, 2017, pp.
  4148--4158.

\bibitem{jacobsen2018excessive}
J{\"o}rn-Henrik Jacobsen, Jens Behrmann, Richard Zemel, and Matthias Bethge,
\newblock ``Excessive invariance causes adversarial vulnerability,''
\newblock {\em arXiv preprint arXiv:1811.00401}, 2018.

\bibitem{biggio2012poisoning}
Battista Biggio, Blaine Nelson, and Pavel Laskov,
\newblock ``Poisoning attacks against support vector machines,''
\newblock {\em arXiv preprint arXiv:1206.6389}, 2012.

\bibitem{shafahi2018poison}
Ali Shafahi, W~Ronny Huang, Mahyar Najibi, Octavian Suciu, Christoph Studer,
  Tudor Dumitras, and Tom Goldstein,
\newblock ``Poison frogs! targeted clean-label poisoning attacks on neural
  networks,''
\newblock in {\em Advances in Neural Information Processing Systems}, 2018, pp.
  6103--6113.

\bibitem{steinhardt2017certified}
Jacob Steinhardt, Pang Wei~W Koh, and Percy~S Liang,
\newblock ``Certified defenses for data poisoning attacks,''
\newblock in {\em Advances in neural information processing systems}, 2017, pp.
  3517--3529.

\bibitem{tanay2018built}
Thomas Tanay, Jerone~TA Andrews, and Lewis~D Griffin,
\newblock ``Built-in vulnerabilities to imperceptible adversarial
  perturbations,''
\newblock {\em arXiv preprint arXiv:1806.07409}, 2018.

\bibitem{munoz2017towards}
Luis Mu{\~n}oz-Gonz{\'a}lez, Battista Biggio, Ambra Demontis, Andrea Paudice,
  Vasin Wongrassamee, Emil~C Lupu, and Fabio Roli,
\newblock ``Towards poisoning of deep learning algorithms with back-gradient
  optimization,''
\newblock in {\em Proceedings of the 10th ACM Workshop on Artificial
  Intelligence and Security}. ACM, 2017, pp. 27--38.

\bibitem{koh2017understanding}
Pang~Wei Koh and Percy Liang,
\newblock ``Understanding black-box predictions via influence functions,''
\newblock in {\em Proceedings of the 34th International Conference on Machine
  Learning-Volume 70}. JMLR. org, 2017, pp. 1885--1894.

\bibitem{hansen2016cma}
Nikolaus Hansen,
\newblock ``The cma evolution strategy: A tutorial,''
\newblock {\em arXiv preprint arXiv:1604.00772}, 2016.

\bibitem{ha2018recurrent}
David Ha and J{\"u}rgen Schmidhuber,
\newblock ``Recurrent world models facilitate policy evolution,''
\newblock in {\em Advances in Neural Information Processing Systems}, 2018, pp.
  2450--2462.

\bibitem{ben2007analysis}
Shai Ben-David, John Blitzer, Koby Crammer, and Fernando Pereira,
\newblock ``Analysis of representations for domain adaptation,''
\newblock in {\em Advances in neural information processing systems}, 2007, pp.
  137--144.

\bibitem{ganin2016domain}
Yaroslav Ganin, Evgeniya Ustinova, Hana Ajakan, Pascal Germain, Hugo
  Larochelle, Fran{\c{c}}ois Laviolette, Mario Marchand, and Victor Lempitsky,
\newblock ``Domain-adversarial training of neural networks,''
\newblock {\em The Journal of Machine Learning Research}, vol. 17, no. 1, pp.
  2096--2030, 2016.

\bibitem{ben2010theory}
Shai Ben-David, John Blitzer, Koby Crammer, Alex Kulesza, Fernando Pereira, and
  Jennifer~Wortman Vaughan,
\newblock ``A theory of learning from different domains,''
\newblock {\em Machine learning}, vol. 79, no. 1-2, pp. 151--175, 2010.

\bibitem{kingma2014adam}
Diederik~P Kingma and Jimmy Ba,
\newblock ``Adam: A method for stochastic optimization,''
\newblock {\em arXiv preprint arXiv:1412.6980}, 2014.

\bibitem{tieleman2017divide}
T~Tieleman and G~Hinton,
\newblock ``Divide the gradient by a running average of its recent magnitude.
  coursera: Neural networks for machine learning,''
\newblock {\em Technical Report.}, 2017.

\bibitem{luo2019adaptive}
Liangchen Luo, Yuanhao Xiong, Yan Liu, and Xu~Sun,
\newblock ``Adaptive gradient methods with dynamic bound of learning rate,''
\newblock {\em arXiv preprint arXiv:1902.09843}, 2019.

\bibitem{hochreiter1997flat}
Sepp Hochreiter and J{\"u}rgen Schmidhuber,
\newblock ``Flat minima,''
\newblock {\em Neural Computation}, vol. 9, no. 1, pp. 1--42, 1997.

\bibitem{goodfellow2014qualitatively}
Ian~J Goodfellow, Oriol Vinyals, and Andrew~M Saxe,
\newblock ``Qualitatively characterizing neural network optimization
  problems,''
\newblock {\em arXiv preprint arXiv:1412.6544}, 2014.

\end{thebibliography}
	
\end{document}